\documentclass[11pt,a4paper]{article}
\PassOptionsToPackage{dvipsnames}{xcolor}
\usepackage[hyperref]{acl2019}
\usepackage{times}
\usepackage{latexsym}

\usepackage{url}

\usepackage{multirow}
\usepackage{amsmath}
\usepackage{capt-of}
\usepackage{tabularx}
\usepackage[caption=false]{subfig}
\usepackage{epsfig}
\usepackage{amssymb}
\usepackage{amsfonts}
\usepackage{booktabs}
\usepackage{scalerel}
\usepackage[inline]{enumitem}
\usepackage{listings}
\usepackage{varwidth}
\usepackage[export]{adjustbox}
\usepackage{tikz}
\usetikzlibrary{tikzmark}
\usepackage{todonotes}
\usepackage{cleveref}

\usepackage{stmaryrd}
\usepackage{bbm}
\usepackage{tablefootnote}
\usepackage{framed}

\newcommand{\tabincell}[2]{\begin{tabular}{@{}#1@{}}#2\end{tabular}}

\newcommand{\stos}{\textsc{Seq2Seq}}
\newcommand{\ptos}{\textsc{Pair2Seq}}
\newcommand{\unansq}{\textsc{UnAnsQ}}
\DeclareMathOperator\softmax{\operatorname{softmax}}

\DeclareMathOperator\sigmoid{\operatorname{sigmoid}}
\DeclareMathOperator\mytanh{\operatorname{tanh}}

\DeclareMathOperator*{\lstm}{\operatorname{f_{LSTM}}}
\DeclareMathOperator*{\bilstm}{\operatorname{f_{BiLSTM}}}

\usepackage{algorithm}
\usepackage[noend]{algpseudocode}

\definecolor{deepblue}{rgb}{0,0,0.5}
\definecolor{officeblue}{RGB}{0,102,204}
\definecolor{deepred}{rgb}{0.6,0,0}
\definecolor{deepgreen}{rgb}{0,0.5,0}
\definecolor{mybrickred}{RGB}{182,50,28}

\definecolor{fillcolor}{RGB}{216,217,252}

\algnewcommand\algorithmicrequireb{{\hspace{0.95cm}}}
\algnewcommand\INPTDESCB{\item[\algorithmicrequireb]}

\algnewcommand\algorithmicfuncdesc{\textbf{Function:}}
\algnewcommand\FUNCDESC{\item[\algorithmicfuncdesc]}
\algnewcommand\algorithmicfuncdescb{{\hspace{0.86cm}}}
\algnewcommand\FUNCDESCB{\item[\algorithmicfuncdescb]}
\algnewcommand{\algorithmicgoto}{\textbf{goto}}
\algnewcommand{\Goto}[1]{\algorithmicgoto~\ref{#1}}

\aclfinalcopy 
\def\aclpaperid{2827} 

\title{Learning to Ask Unanswerable Questions \\ for Machine Reading Comprehension}

\author{Haichao Zhu$^\dag$\thanks{\ \  Contribution during internship at Microsoft Research Asia.},~~Li Dong$^\ddag$,~~Furu Wei$^\ddag$,~~Wenhui Wang$^\ddag$,~~Bing Qin$^{\dag\natural}$,~~Ting Liu$^{\dag\natural}$\\
  $^\dag$Harbin Institute of Technology, Harbin, China \\
  $^\ddag$Microsoft Research, Beijing, China \\
  $^\natural$Peng Cheng Laboratory, Shenzhen, China \\
  \texttt{\{hczhu,qinb,tliu\}@ir.hit.edu.cn} \\
  \texttt{\{lidong1,fuwei,wenwan\}@microsoft.com} \\}

\date{}

\begin{document}
\maketitle
\begin{abstract}
Machine reading comprehension with unanswerable questions is a challenging task.
In this work, we propose a data augmentation technique by automatically generating relevant unanswerable questions according to an answerable question paired with its corresponding paragraph that contains the answer.
We introduce a pair-to-sequence model for unanswerable question generation, which effectively captures the interactions between the question and the paragraph.
We also present a way to construct training data for our question generation models by leveraging the existing reading comprehension dataset.
Experimental results show that the pair-to-sequence model performs consistently better compared with the sequence-to-sequence baseline.
We further use the automatically generated unanswerable questions as a means of data augmentation on the SQuAD 2.0 dataset, yielding $1.9$ absolute F1 improvement with BERT-base model and $1.7$ absolute F1 improvement with BERT-large model.
\end{abstract}

\section{Introduction}

Extractive reading comprehension~\cite{deepmindqa,squad} obtains great attentions from both research and industry in recent years. End-to-end neural models~\cite{bidaf,rnet,qanet} have achieved remarkable performance on the task if answers are assumed to be in the given paragraph.
Nonetheless, the current systems are still not good at deciding whether no answer is presented in the context~\cite{squad2}. For unanswerable questions, the systems are supposed to abstain from answering rather than making unreliable guesses, which is an embodiment of language understanding ability.

We attack the problem by automatically generating unanswerable questions for data augmentation to improve question answering models.
The generated unanswerable questions should not be too easy for the question answering model so that data augmentation can better help the model. For example, a simple baseline method is randomly choosing a question asked for another paragraph, and using it as an unanswerable question. However, it would be trivial to determine whether the retrieved question is answerable by using word-overlap heuristics, because the question is irrelevant to the context~\cite{qalexical}.
In this work, we propose to generate unanswerable questions by editing an answerable question and conditioning on the corresponding paragraph that contains the answer. So the generated unanswerable questions are more lexically similar and relevant to the context. Moreover, by using the answerable question as a prototype and its answer span as a plausible answer, the generated examples can provide more discriminative training signal to the question answering model.

\begin{figure}[t]
\begin{framed}
\footnotesize
\textbf{Title:} Victoria (Australia)\\
\textbf{Paragraph:} 
\dots Public schools, also known as state or government schools, are funded and run directly by the {\color{Bittersweet}Victoria Department of Education} . Students do not pay tuition fees, but some extra costs are levied. Private fee-paying schools include parish schools
\dots
\vspace{0.08in} \\
\textbf{Ans. Question}: What organization runs \textit{the public schools} in Victoria?  \\
\textbf{UnAns. Question}: What organization runs \textit{the waste management} in Victoria? \vspace{0.08in}  \\
\textbf{(Plausible) Answer}: {\color{Bittersweet}Victoria Department of Education}
\end{framed}
\caption{An example taken from the SQuAD 2.0 dataset. The annotated (plausible) answer span in the paragraph is used as a pivot to align the pair of answerable and unanswerable questions.}
\label{fig:squad2_example}
\end{figure}

To create training data for unanswerable question generation, we use (plausible) answer spans in paragraphs as pivots to align pairs of answerable questions and unanswerable questions.
As shown in Figure~\ref{fig:squad2_example}, the answerable and unanswerable questions of a paragraph are aligned through the text span ``\textit{Victoria Department of Education}'' for being both the answer and plausible answer.
These two questions are lexically similar and both asked with the same answer type in mind.
In this way, we obtain the data with which the models can learn to ask unanswerable questions by editing answerable ones with word exchanges, negations, etc.
Consequently, we can generate a mass of unanswerable questions with existing large-scale machine reading comprehension datasets.

Inspired by the neural reading comprehension models~\cite{dcn,huang2018fusionnet}, we introduce a pair-to-sequence model to better capture the interactions between questions and paragraphs.
The proposed model first encodes input question and paragraph separately, and then conducts attention-based matching to make them aware of each other. Finally, the context-aware representations are used to generate outputs.
To facilitate the use of context words during the generation process, we also incorporate the copy mechanism~\cite{copynet, copy-summarization}.

Experimental results on the unanswerable question generation task shows that the pair-to-sequence model generates consistently better results over the  sequence-to-sequence baseline and performs better with long paragraphs than with short answer sentences.
Further experimental results show that the generated unanswerable questions can improve multiple machine reading comprehension models.
Even using BERT fine-tuning as a strong reading comprehension model, we can still obtain a $1.9$\% absolute improvement of F1 score with BERT-base model and $1.7$\% absolute F1 improvement with BERT-large model.

\section{Related Work}
\textbf{Machine Reading Comprehension (MRC)}
Various large-scale datasets~\cite{deepmindqa,squad,marco,triviaQA,squad2,narrativeqa} have spurred rapid progress on machine reading comprehension in recent years.
SQuAD~\cite{squad} is an extractive benchmark whose questions and answers spans are annotated by humans.
Neural reading comprehension systems  ~\cite{match-lstm,bidaf,rnet,reinforceijcai18hu,huang2018fusionnet,san,qanet,multigranu} have outperformed humans on this task in terms of automatic metrics.
The SQuAD~2.0 dataset~\cite{squad2} extends SQuAD with more than $50,000$ crowdsourced unanswerable questions.
So far, neural reading comprehension models still fall behind humans on SQuAD~2.0.
Abstaining from answering when no answer can be inferred from the given document  does require more understanding than barely extracting an answer.

\textbf{Question Generation for MRC}
In recent years, there has been an increasing interest in generating questions for reading comprehension.
~\citet{qgacl17xinyadu} show that neural models based on the encoder-decoder framework can generate significantly better questions than rule-based systems~\cite{qg_rule}.
To generate answer-focused questions, one can simply indicate the answer positions in the context with extra features~\cite{qgacl17rl,qgnlpcc18qyz,qgacl18xinyadu,qgemnlp18baidu,unilm19}.
\citet{qg_naacl18} and \citet{qgaaai19} separate answer representations for further matching.
\citet{qgijcai18} introduce a latent variable for capturing variability and an observed variable for controlling question types.
In summary, the above mentioned systems aim to generate answerable questions with certain context.
On the contrary, our goal is to generate unanswerable questions.

\textbf{Adversarial Examples for MRC}
To evaluate the language understanding ability of pre-trained systems, \citet{adversarialemnl17} construct adversarial examples by adding distractor sentences that do not contradict question answering for humans to the paragraph.
\citet{simpleandeffective} and \citet{qgnlpcc18cqt} use questions to retrieve paragraphs that do not contain the answer as adversarial examples.
\citet{squad2} create unanswerable questions through rigid rules, which swap entities, numbers and antonyms of answerable questions.
It has been shown that adversarial examples generated by rule-based systems are much easier to detect than ones in the SQuAD 2.0 dataset.

\textbf{Data Augmentation for MRC}
Several attempts have been made to augment training data for machine reading comprehension.
We categorize these work according to the type of the augmentation data: external data source, paragraphs or questions.
\citet{bert} fine-tune BERT on the SQuAD dataset jointly with another dataset  TriviaQA~\cite{triviaQA}.
\citet{qanet} paraphrase paragraphs with backtranslation.
Another line of work adheres to generate answerable questions.
\citet{aug_qg} propose to generate questions based on the unlabeled text for semi-supervised question answering.
\citet{aug_race} propose a rule-based system to generate multiple-choice questions with candidate options upon the paragraphs.
We aim at generating unanswerable questions as a means of data augmentation.

\section{Problem Formulation}
Given an answerable question $q$ and its corresponding paragraph $p$ that contains the answer $a$, we aim to generate unanswerable questions $\tilde{q}$ that fulfills certain requirements.
First, it cannot be answered by paragraph $p$.
Second, it must be relevant to both answerable question $q$ and paragraph $p$, which refrains from producing irrelevant questions.
Third, it should ask for something of the same type as answer $a$.

\begin{figure}
    \centering
    \resizebox{\textwidth/2-0.5cm}{!}{
    \includegraphics{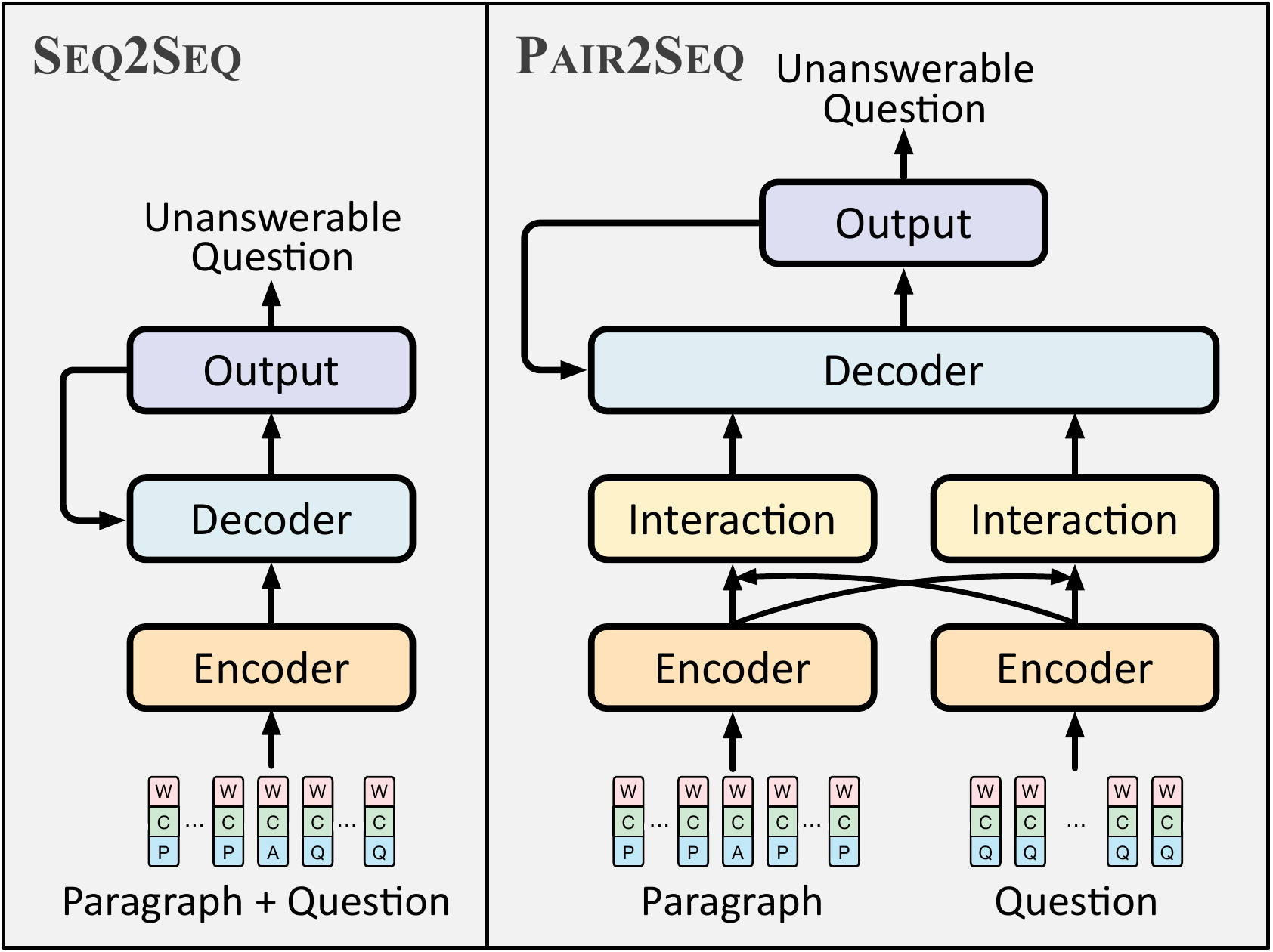}
    }
    \caption{Diagram of the proposed pair-to-sequence model and sequence-to-sequence model.
    The input embeddings is the sum of the word embeddings, the character embeddings and the token type embeddings.
    The input questions are all answerable.}
    \label{fig:model}
\end{figure}

As shown in Figure~\ref{fig:model}, we investigate two simple neural models built upon encoder-decoder architecture~\cite{encoder-decoder:14,encoder-decoder:15} to generate unanswerable questions. 
A sequence-to-sequence model takes the concatenated paragraph and question as input, and encodes the input in a sequential manner.
A pair-to-sequence model is further introduced to capture the interactions between inputs.
The decoder of two models generates unanswerable questions sequentially.
We factorize the probability of generating the unanswerable question $P(\tilde{q}|q,p,a)$ as:
\begin{align}
    P(\tilde{q}|q,p,a)=\prod_{t=1}^{|\tilde{q}|}P(\tilde{q}_t|\tilde{q}_{<t},q,p,a)
\end{align}
where $\tilde{q}_{<t}=\tilde{q}_1 \dots \tilde{q}_{t-1}$.

\subsection{Sequence-to-Sequence Model}
In the sequence-to-sequence model, paragraph and question pairs are packed into an ordered sequence $x$ with a special separator in between.
To indicate answers in paragraphs, we introduce token type embeddings which can also be used to distinguish questions from paragraphs in sequence-to-sequence model.
As we can see in Figure~\ref{fig:model}, the token type can be answer (A), paragraph (P), or question (Q).
For a given token, we construct the input representation $\mathbf{e}_i$ by summing the corresponding word embeddings, character embeddings and token type embeddings.
Here characters are embedded by an embedding matrix followed by a max pooling layer.

We apply a single-layer bi-directional recurrent neural networks with long short-term memory units (LSTM;~\citealp{lstm}) to produce encoder hidden states $\mathbf{h}_i=\bilstm(\mathbf{h}_{i-1}, \mathbf{e}_i)$.
On each decoding step $t$, the hidden states of decoder (a single-layer unidirectional LSTM network) are computed by $\mathbf{s}_t=\lstm(\mathbf{s}_{t-1}, [\mathbf{y}_{t-1}; \mathbf{c}_{t-1}])$, where $\mathbf{y}_{t-1}$ is the word embedding of previously predicted token and $\mathbf{c}_{t-1}$ is the encoder context vector of previous step. Besides, we use an attention mechanism to summarize the encoder-side information into $\mathbf{c}_{t}$ for current step. The attention distribution $\gamma_t$ over source words is computed as in~\citet{global-attn}:
\begin{align}
score(\mathbf{h}_i , \mathbf{s}_t)&=\mathbf{h}_i^{\mathrm{T}}\mathbf{W}_{\gamma}\mathbf{s}_t\\
\gamma_{i,t}&={\exp(score(\mathbf{h}_i,\mathbf{s}_t))} / Z_t \label{eq:attn:score} \\
\mathbf{c}_t&=\sum_{i}^{|x|}{\gamma_{i,t} \mathbf{h}_i}\label{eq:context}
\end{align}
where $Z_t = {\sum_{k}^{|x|}\exp(score(\mathbf{h}_k,\mathbf{s}_t))}$,  $\mathbf{W}_\gamma$ in score function is a learnable parameter.

Next, $\mathbf{s}_t$ is concatenated with $\mathbf{c}_t$ to produce the vocabulary distribution $P_{v}$:
\begin{equation}
    P_{v}=\softmax(\mathbf{W}_v[\mathbf{s}_t;\mathbf{c}_t] + \mathbf{b}_{v})
\end{equation}
where $\mathbf{W}_v$ and $\mathbf{b}_{v}$ are learnable parameters.
Copy mechanism~\cite{copy-summarization} is incorporated to directly copy words from inputs, because words in paragraphs or source questions are of great value for unanswerable question generation.
Specifically, we use $\mathbf{s}_t$ and $\mathbf{c}_t$ to produce a gating probability $g_t$:
\begin{equation}
g_t=\sigmoid(\mathbf{W}_g[\mathbf{s}_t;\mathbf{c}_t] + \mathbf{b}_{g})
\end{equation}
where $\mathbf{W}_g$ and $\mathbf{b}_{g}$ are learnable parameters.
The gate $g_t$ determines whether generating a word from the vocabulary or copying a word from inputs.
Finally, we obtain the probability of generating $\tilde{q}_t$ by:
\begin{equation}
P(\tilde{q}_t|\tilde{q}_{<t},q,p,a)=g_t P_{v}(\tilde{q}_t) + (1-g_t)\sum_{i \in \zeta_{\tilde{q}_t}}\hat{\gamma}_{i,t} \nonumber
\end{equation}
where $\zeta_{\tilde{q}_t}$ denotes all the occurrence of $\tilde{q}_t$ in inputs, and the copying score $\hat{\gamma}_t$ is computed in the same way as attention scores $\gamma_t$ (see Equation~\eqref{eq:attn:score}) while using different parameters.

\subsection{Pair-to-Sequence Model}
Paragraph and question interactions play a vitally important role in machine reading comprehension. 
The interactions make the paragraph and question aware of each other and help to predict the answer more precisely.
Therefore we propose a pair-to-sequence model, conducting attention based interactions in encoder and subsequently decoding with two series of representations.

In pair-to-sequence model, the paragraph and question are embedded as in sequence-to-sequence model, but encoded separately by weight-shared bi-directional LSTM networks, yielding $\mathbf{h}_i^p=\bilstm(\mathbf{h}_{i-1}^p, \mathbf{e}_{i-1}^p)$ as paragraph encodings and $\mathbf{h}_i^q=\bilstm(\mathbf{h}_{i-1}^q, \mathbf{e}_{i-1}^q)$ as question encodings.
The same attention mechanism as in sequence-to-sequence model is used in the following interaction layer to produce question-aware paragraph representations $\Tilde{\mathbf{h}}_i^p$:
\begin{align}
\alpha_{i,j}&=\exp(score(\mathbf{h}_i^p,\mathbf{h}_j^q))/Z_i\\
\hat{\mathbf{h}}_i^p&=\sum_{j=1}^{|q|}\alpha_{i,j}\mathbf{h}_j^q\\
\Tilde{\mathbf{h}}_i^p&=\mytanh (\mathbf{W}_p[\mathbf{h}_i^p;\hat{\mathbf{h}}_i^p] + \mathbf{b}_p)
\end{align}
where $Z_i=\sum_{k=1}^{|q|}\exp(score(\mathbf{h}_i^p,\mathbf{h}_k^q))$ ,$\mathbf{W}_p$ and $\mathbf{b}_p$ are learnable parameters. Similarly, the paragraph-aware question representations $\Tilde{\mathbf{h}}_i^q$ are produced by:
\begin{align}
    \beta_{i,j}&={\exp(score(\mathbf{h}_i^p,\mathbf{h}_j^q))}/{Z_j}\\
    \hat{\mathbf{h}}_i^q&=\sum_{i=1}^{|p|}\beta_{i,j}\mathbf{h}_i^p\\
    \Tilde{\mathbf{h}}_j^q&=\mytanh (\mathbf{W}_q[\mathbf{h}_j^q;\hat{\mathbf{h}}_j^q] + \mathbf{b}_q)
\end{align}
where $Z_j=\sum_{k=1}^{|p|}\exp(score(\mathbf{h}_k^p,\mathbf{h}_j^q))$, $\mathbf{W}_q$ and $\mathbf{b}_q$ are learnable parameters. 

Accordingly, the decoder now takes paragraph context $\mathbf{c}^p_{t-1}$ and question context $\mathbf{c}^q_{t-1}$ as encoder context, computed as $\mathbf{c}_t$ (see Equation \eqref{eq:context}) in sequence-to-sequence model, to update decoder hidden states $\mathbf{s}_t=\lstm(\mathbf{s}_{t-1},[\mathbf{y}_{t-1};\mathbf{c}^p_{t-1};\mathbf{c}^q_{t-1}])$ and predict tokens. Copy mechanism is also adopted as described before, and copying words from both the paragraph and question is viable.

\subsection{Training and Inference}
The training objective is to minimize the negative likelihood of the aligned unanswerable question $\tilde{q}$ given the answerable question $q$ and its corresponding paragraph $p$ that contains the answer $a$:
\begin{align}
\mathcal{L}&=-\sum_{(\tilde{q},q,p,a)\in\mathcal{D}}\log P(\tilde{q}|q,p,a;\theta)
\end{align}
where $\mathcal{D}$ is the training corpus and $\theta$ denotes all the parameters. Sequence-to-sequence and pair-to-sequence models are trained with the same objective.

During inference, the unanswerable question for question answering pair $(q,p,a)$ is obtained via $\textrm{argmax}_{q'}P(q'|q,p,a)$, where $q'$ represents candidate outputs.
Beam search is used to avoid iterating over all possible outputs.

\section{Experiments}
We conduct experiments on the SQuAD 2.0 dataset~\cite{squad2}.
The extractive machine reading benchmark contains about $100,000$ answerable questions and over $50,000$ crowdsourced unanswerable questions towards Wikipedia paragraphs.
Crowdworkers are requested to craft unanswerable questions that are relevant to the given paragraph. Moreover, for each unanswerable question, a plausible answer span is annotated, which indicates the incorrect answer obtained by only relying on type-matching heuristics.
Both answers and plausible answers are text spans in the paragraphs.

\subsection{Unanswerable Question Generation}

\subsubsection{Training Data Construction}

We use (plausible) answer spans in paragraphs as pivots to align pairs of answerable questions and unanswerable questions.
An aligned pair is shown in Figure~\ref{fig:squad2_example}.
As to the spans that correspond to multiple answerable and unanswerable questions, we sort the pairs by Levenshtein distance~\cite{levenshtein1966} and keep the pair with the minimum distance, and make sure that each question is only paired once.

We obtain $20,240$ aligned pairs from the SQuAD 2.0 dataset in total.
The Levenshtein distance between the answerable and unanswerable questions in pairs is $3.5$ on average.
Specifically, the $17,475$ pairs extracted from the SQuAD 2.0 training set are used to train generation models.
Since the SQuAD 2.0 test set is hidden, we randomly sample $46$ articles from the SQuAD 2.0 training set with $1,805$~($\sim${$10$\%}) pairs as holdout set and evaluate generation models with $2,765$ pairs extracted the SQuAD 2.0 development set.

\subsubsection{Settings}
We implement generation models upon OpenNMT~\cite{opennmt}.
We preprocess the corpus with the spaCy toolkit for tokenization and sentence segmentation.
We lowercase tokens and build the vocabulary on SQuAD 2.0 training set with word frequency threshold of $9$ to remove most noisy tokens introduced in data collection and tokenization.
We set word, character and token type embeddings dimension to $300$.
We use the \verb|glove.840B.300d| pre-trained embeddings~\cite{glove} to initialize word embeddings, and do further updates during training.
Both encoder and decoder share the same vocabulary and word embeddings.
The hidden state size of LSTM network is $150$.
Dropout probability is set to $0.2$.
The data are shuffled and split into mini-batches of size $32$ for training.
The model is optimized with Adagrad~\cite{adagrad} with an initial learning rate of $0.15$. During inference, the beam size is $5$.
We prohibit producing unknown words by setting the score of \verb|<unk>| token to \verb|-inf|.
We filter the beam outputs that make no differences to the input question.

\subsubsection{Evaluation Metrics}
The generation quality is evaluated using three automatic evaluation metrics: BLEU~\cite{bleu}, ROUGE~\cite{rouge} and GLEU~\cite{GLEU}. 
BLEU\footnote{\url{github.com/moses-smt/mosesdecoder}} is a commonly used metric in machine translation that computes n-gram precisions over references. 
Recall-oriented ROUGE\footnote{\url{pypi.org/project/pyrouge}} metric is widely adopted in summarization, and ROUGE-L measures longest common subsequence between system outputs and references. 
GLEU\footnote{\url{github.com/cnap/gec-ranking}} is a variant of BLEU with the modification that penalizes system output n-grams that present in input but absent from the reference.
This makes GLEU a preferable metric for tasks with subtle but critical differences in a monolingual setting as in our unanswerable question generation task.

We also conduct human evaluation on $100$ samples in three criteria:
(1) unanswerability, which indicates whether the question is unanswerable or not; 
(2) relatedness, which measures semantic relatedness between the generated question and input question answering pair;
(3) readability, which indicates the grammaticality and fluency.
We ask three raters to score the generated questions in terms of relatedness and readability on a $1$-$3$ scale~($3$ for the best) and determine the answerability in binary~($1$ for unanswerable).
The raters are not aware of the question generation methods in advance.

\subsubsection{Results} 

\begin{table*}[!htp]
    \centering
    \resizebox{\textwidth}{!}{
    \begin{tabular}{lcc|cc|ccc}
    \toprule
    Model & GLEU-3 & GLEU-4 & BLEU-3 & BLEU-4 & ROUGE-2 & ROUGE-3 & ROUGE-L \\
    \midrule
    \stos & 33.13 & 27.39 & 36.80 & 27.84 & 46.54 & 32.98 & 64.28 \\
    \ptos & \textbf{35.06} & \textbf{29.43} & \textbf{37.67} & \textbf{29.17} & \textbf{47.46} & \textbf{34.18} & \textbf{65.24} \\
    \quad- Paragraph (+AS) & 34.42 & 28.43 & 37.35 & 28.44 & 47.13 & 33.29 & 65.02\\
    \quad- Paragraph & 33.58 & 27.54 & 35.89 & 26.99 & 46.14 & 31.45 & 64.78 \\
    \quad- Question & 9.40 & 6.21 & 6.7 & 3.1 & 12.64 & 5.63 & 32.26 \\
    \quad- Copy & 25.06 & 19.80 & 36.06 & 22.84 & 33.40 & 20.45 & 52.76\\
    \bottomrule
    \end{tabular}}
    \caption{Automatic evaluation results. Higher score is better and the best performance for each evaluation metric is highlighted in \textbf{boldface}. ``- Paragraph (+AS)'' represents replacing paragraphs with answer sentences.}
    \label{tab:qg_results}
\end{table*}

\begin{table}[!t]
    \centering
    \begin{tabular}{lcc}
         \toprule
         ~ & EM / F1 & $\triangle$ \\
         \midrule
         BNA &  59.7/62.7 & - \\
         \quad+ \unansq\ & 61.0/63.5 & +1.3/+0.8 \\
         \midrule
         DocQA &  61.9/64.5 & - \\
         \quad+ \unansq\ & 62.4/65.3 & +0.5/+0.8 \\
         \midrule
         BERT$_{Base}$ &  74.3/77.4 & - \\
         \quad+ \unansq\ & 76.4/79.3 & +2.1/+1.9 \\
         \midrule
         BERT$_{Large}$ &  78.2/81.3 & - \\
         \quad+ \unansq\ & 80.0/83.0 & +1.8/+1.7 \\
         \bottomrule
    \end{tabular}
    \caption{Experimental results of applying data augmentation to reading comprehension models on the SQuAD 2.0 dataset. ``$\triangle$'' indicates absolute improvement.}
    \label{tab:qa_model_ablation}
\end{table}

Results of the automatic evaluation are shown in Table~\ref{tab:qg_results}.
We find that the proposed pair-to-sequence model that captures interactions between paragraph and question performs consistently better than sequence-to-sequence model.
Moreover, replacing the input paragraph with the answer sentence hurts model performance, which indicates that using the whole paragraph as context provides more helpful information to unanswerable question generation.
We also try to generate unanswerable questions by only relying on answerable questions (see ``-Paragraph''), or the paragraph (see ``-Question''). Unsurprisingly, both ablation models obtain worse performance compared with the full model. These two ablation results also demonstrate that the input answerable question helps more to improve performance compared with the input paragraph. We argue that the original answerable question provides more direct information due to the fact that the average edit distance between the example pairs is $3.5$.
At last, we remove the copy mechanism that restrains prediction tokens to the vocabulary. 
The results indicate the necessity of copying tokens from answerable questions and paragraphs to outputs, which relieves the out-of-vocabulary problem.

\begin{table}[t]
\centering
\begin{tabular}{lccc}
\toprule
& \textsc{UnAns} & \textsc{Rela} & \textsc{Read} \\
\midrule
\textsc{TfIdf}   & \textbf{0.96} & 1.52 & \textbf{2.98} \\
\stos  & 0.62 & 2.88 & 2.39 \\
\ptos & 0.65 & \textbf{2.95} & 2.61 \\
\midrule
Human    & 0.95 & 2.96 & 3 \\
 \bottomrule
\end{tabular}
\caption{Human evaluation results.
Unanswerability (\textsc{UnAns}): 1 for unanswerable, 0 otherwise.
Relatedness (\textsc{Rela}): 3 for relevant to both answerable question and paragraph, 2 for relevant to only one, 1 for irrelevant. Readability (\textsc{Read}): 3 for fluent, 2 for minor grammatical errors, 1 for incomprehensible.
}
\label{tab:human_evaluation}
\end{table}

\begin{table}[t]
    \centering
    \begin{tabular}{lrrr}
         \toprule
         Type & S2S & P2S & Human  \\
         \midrule
         Negation & 42\% & 54\% & 32\% \\
         Antonym & 4\% & 5\% & 9\% \\
         \tabincell{l}{Entity Swap}& 17\% & 20\% & 20\% \\
         \tabincell{l}{Mutual Exclusion} & 2\% & 0\% & 12\% \\
         \tabincell{l}{Impossible Condition} & 8\% & 12\% & 25\% \\
         \tabincell{l}{Other} & 27\% & 8\% & 2\% \\
         \bottomrule
    \end{tabular}
    \caption{Types of unanswerable questions generated by models and humans, we refer the reader to~\cite{squad2} for detail definition of each type. ``S2S'' represents the sequence-to-sequence baseline and ``P2S'' is our proposed pair-to-sequence model.}
    \label{tab:unanswer_type}
\end{table}

Table~\ref{tab:human_evaluation} shows the human evaluation results of generated unanswerable questions.
We compare with the baseline method \textsc{TfIdf}, which uses the input answerable question to retrieve similar questions towards other articles as outputs.\label{tfidf_desc}
The retrieved questions are mostly unanswerable and readable, but they are not quite relevant to the question answering pair. Notice that being relevant is demonstrated to be important for data augmentation in further experiments on machine reading comprehension.
Here pair-to-sequence model still outperforms sequence-to-sequence model in terms of all three metrics.
But the differences in human evaluation are not as notable as in the automatic metrics.

\begin{figure*}[t]
    \begin{framed}
    \footnotesize
         \textbf{Title:} Victoria (Australia)\\
         \textbf{Paragraph:} 
         Victorian schools are either publicly or privately funded. Public schools, also known as {\color{blue}state or government schools}, are funded and run directly by the {\color{Bittersweet}Victoria Department of Education} . Students do not pay tuition fees, but {\color{red}some extra costs} are levied. Private fee-paying schools include parish schools run by the {\color{ForestGreen}Roman Catholic Church} and independent schools similar to British public schools. Independent schools are usually affiliated with Protestant churches. Victoria also has several private Jewish and Islamic primary and secondary schools. Private schools also receive some 
         ...
         \vspace{0.08in}
         \\
         \textbf{Question:} What organization runs the public schools in Victoria?  \\
         \textbf{Human:} What organization runs the waste management in Victoria?  \\
         \textbf{\stos:} what organization runs the public schools in \underline{texas} ? \\
         \textbf{\ptos:} what organization \underline{never} runs the public schools in victoria ?  \\
         \vspace{0.08in}
         \textbf{(Plausible) Answer:} {\color{Bittersweet}Victoria Department of Education}  \\
         \textbf{Question:} What church runs some private schools in Victoria? \\
         \textbf{Human:} What church runs public schoolsin Victoria?  \\
         \textbf{\stos:} what church runs some private schools ?  \\
         \textbf{\ptos:} what church \underline{no longer} runs some private schools in victoria ?  \\
         \vspace{0.08in}
         \textbf{(Plausible) Answer:} {\color{ForestGreen}Roman Catholic Church} \\
         \textbf{Question:} Since students do not pay tuition, what do they have to pay for schooling in Victoria?  \\
         \textbf{Human:} What is covered by the state in addition to tuition?  \\
         \textbf{\stos:} since students do not \underline{pay to pay} schooling in victoria ? \\
         \textbf{\ptos:} since students do \underline{n't} pay tuition , what do they have to pay for schooling in victoria ?  \\
         \vspace{0.08in}
         \textbf{(Plausible) Answer:} {\color{red}some extra costs}  \\
         \textbf{Question:} What are public schools in Victoria?  \\
         \textbf{Human:} What are public banks in Victoria?  \\
         \textbf{\stos:} what are \underline{n't} public schools in victoria ? \\
         \textbf{\ptos:} what are public schools \underline{not} in victoria ?  \\
         \textbf{(Plausible) Answer:} {\color{blue}state or government schools}
    \end{framed}
    \caption{Sample output generated by human, sequence-to-sequence model, and pair-to-sequence model.
    The (plausible) answer span of questions are marked in colors and main difference of model outputs are underlined.}
    \label{tab:examples}
\end{figure*}

As shown in Table~\ref{tab:unanswer_type}, we further randomly sample $100$ system outputs to analyze the types of generated unanswerable questions.
We borrow the types defined in~\citet{squad2} for SQuAD 2.0. We categorize the outputs with grammatical errors that make them hard to understand into \verb|Other|. Samples that fall into \verb|Impossible Condition| are mainly produced by non-entity substitution.
We can see that models tend to generate unanswerable questions by inserting negation and swapping entities. These two types are also most commonly used when crowdworkers pose unanswerable questions according to answerable ones.
We also find that the current models still have difficulties in utilizing antonyms and exclusion conditions, which could be improved by incorporating external resources.

In Figure~\ref{tab:examples}, we present a sample paragraph and its corresponding answerable questions and generated unanswerable questions.
In the first example, two models generate unanswerable questions by swapping the location entity ``\textit{Victoria}'' with ``\textit{texas}'' and inserting negation word ``\textit{never}'', respectively.
In the second example, sequence-to-sequence model omits the condition ``\textit{in Victoria}'' and yields an answerable question.
Pair-to-sequence model inserts the negation ``\textit{no longer}'' properly, which is not mentioned in the paragraph.
In the third example, grammatical errors are found in the output of \stos.
The last example shows that inserting negation words in different positions (``\textit{n't public}'' versus ``\textit{not in victoria}'') can express different meanings.
Such cases are critical for generated questions' answerability, which is hard to handle in a rule-based system.

\subsection{Data Augmentation for Machine Reading Comprehension}

\subsubsection{Question Answering Models}
We apply our automatically generated unanswerable questions as augmentation data to the following reading comprehension models:

\paragraph{BiDAF-No-Answer (BNA)}
BiDAF~\cite{bidaf} is a benchmark model on extractive machine reading comprehension. Based on BiDAF, \citet{bna} propose the BiDAF-No-Answer model to predict the distribution of answer candidates and the probability of a question being unanswerable at the same time.

\paragraph{DocQA}
\citet{simpleandeffective} propose the DocQA model to address document-level reading comprehension. The no-answer probability is also predicted jointly.

\paragraph{BERT Fine-Tuning}
It is the state-of-the-art model on unanswerable machine reading comprehension.
We adopt the uncased version of BERT~\cite{bert} for fine-tuning.
The batch sizes of BERT-base and BERT-large are set to $12$ and $24$ respectively.
The rest hyperparameters are kept untouched as in the official instructions of fine-tuning BERT-Large on SQuAD 2.0.

\subsubsection{Data Augmentation Setup} 
We first generate unanswerable questions using the trained generation model.
Specifically, we use the answerable questions in the SQuAD 2.0 training set, besides ones aligned before, to generate unanswerable questions.
Then we use the paragraph and answers of answerable questions along with the generated questions to construct training examples.
At last, we have an augmentation data containing $69,090$ unanswerable examples. 

We train question answering models with augmentation data in two separate phases.
In the first phase, we train the models by combining the augmentation data and all $86,821$ SQuAD 2.0 answerable examples.
Subsequently, we use the original SQuAD 2.0 training data alone to further fine-tune model parameters.

\subsubsection{Results} 

Exact Match (EM) and F1 are two metrics used to evaluate model performance.
EM measures the percentage of predictions that match ground truth answers exactly.
F1 measures the word overlap between the prediction and ground truth answers.
We use pair-to-sequence model with answerable questions and paragraphs for data augmentation by default.

Table~\ref{tab:qa_model_ablation} shows the exact match and F1 scores of multiple reading comprehension models with and without data augmentation.
We can see that the generated unanswerable questions can improve both specifically designed reading comprehension models and strong BERT fine-tuning models, yielding $1.9$ absolute F1 improvement with BERT-base model and $1.7$ absolute F1 improvement with BERT-large model.
Our submitted model obtains an EM score of $80.75$ and an F1 score of $83.85$ on the hidden test set.

\begin{table}[!t]
    \centering
    \begin{tabular}{lcc}
         \toprule
         ~ & EM / F1 & $\triangle$ \\
         \midrule
         BERT$_{Base}$ &  74.3/77.4 & - \\
         \quad+ \textsc{TfIdf} & 75.0/77.8 & +0.7/+0.4 \\
         \quad+ \textsc{Rule} & 75.6/78.5 & +1.3/+1.1\\
         \quad+ \stos & 75.5/78.2 & +1.2/+0.8 \\
         \quad+ \ptos & 76.4/79.3 & +2.1/+1.9 \\
         \bottomrule
    \end{tabular}
    \caption{Results using different generation methods for data augmentation. ``$\triangle$'' indicates absolute improvement.}
    \label{tab:qa_method_ablation}
\end{table}

As shown in Table~\ref{tab:qa_method_ablation}, pair-to-sequence model proves to be a better option for generating augmentation data than other three methods.
Besides the sequence-to-sequence model, we use answerable questions to retrieve questions from other articles with \textsc{TfIdf}.
The retrieved questions are of little help to improve the model, because they are less relevant to the paragraph as shown in Table~\ref{tab:human_evaluation}.
We refer to the rule-based method~\cite{adversarialemnl17} that swaps entities and replaces words with antonyms as \textsc{Rule}.
In comparison to the above methods, pair-to-sequence model can yield the largest improvement.

\begin{table}[!t]
    \centering
    \begin{tabular}{lcc}
         \toprule
         ~ & EM / F1 & $\triangle$ \\
         \midrule
         BERT$_{Base}$ &  74.3/77.4 & - \\
         \quad+ \unansq $\times$1 & 76.4/79.3 & +2.1/+1.9 \\
         \quad+ \unansq $\times$2 & 76.4/79.4 & +2.1/+2.0 \\
         \quad+ \unansq $\times$3 & 76.6/79.6 & +2.3/+2.2 \\
         \midrule
         BERT$_{Large}$ &  78.2/81.3 & - \\
         \quad+ \unansq$\times$1 & 80.0/83.0 & +1.8/+1.7 \\
         \quad+ \unansq$\times$2 & 80.0/82.9 & +1.8/+1.6 \\
         \quad+ \unansq$\times$3 & 80.1/83.1 & +1.9/+1.8 \\
         \bottomrule
    \end{tabular}
    \caption{Ablation over the size of data augmentation. ``$\times$ N'' means the original size is enhanced  N times. ``$\triangle$'' indicates absolute improvement.}
    \label{tab:qa_size_ablation}
\end{table}

Results in Table~\ref{tab:qa_size_ablation} show that enlarging the size of augmentation data can further improve model performance, especially with the BERT-base model.
We conduct experiments using two and three times the size of the base augmentation data~(i.e., $69,090$ unanswerable questions).
We generate multiple unanswerable questions for each answerable question by using beam search.
Because we only generate unanswerable questions, the data imbalance problem could mitigate the improvement of incorporating more augmentation data.

\section{Conclusions}
In this paper, we propose to generate unanswerable questions as a means of data augmentation for machine reading comprehension.
We produce relevant unanswerable questions by editing answerable questions and conditioning on the corresponding paragraph.
A pair-to-sequence model is introduced in order to capture the interactions between question and paragraph.
We also present a way to construct training data for unanswerable question generation models.
Both automatic and human evaluations show that the proposed model consistently outperforms the sequence-to-sequence baseline.
The results on the SQuAD 2.0 dataset show that our generated unanswerable questions can help to improve multiple reading comprehension models.
As for future work, we would like to enhance the ability to utilize antonyms for unanswerable question generation by leveraging external resources.

\section*{Acknowledgments}
We thank anonymous reviewers for their helpful comments. Qin and Liu were supported by National Natural Science Foundation of China (NSFC) via grants 61632011 and 61772156.

\bibliography{acl2019}
\bibliographystyle{acl_natbib}

\end{document}